\documentclass[letterpaper, 10 pt, conference]{./misc/ieeeconf}  

\usepackage{hyperref}
\usepackage{cite}
\usepackage{amsmath,amssymb,amsfonts}
\usepackage{tikz}
\usetikzlibrary{arrows.meta,calc,positioning}
\usepackage{subcaption}
\usepackage{booktabs}
\usepackage{multicol}
\usepackage{multirow}
\usepackage{threeparttable}
\usepackage[ruled,vlined,linesnumbered]{algorithm2e} 
\usepackage{xcolor}

\IEEEoverridecommandlockouts                              
\title{\LARGE \bf FAR-AVIO: Fast and Robust Schur-Complement Based Acoustic-Visual-Inertial Fusion Odometry with Sensor Calibration
}

\author{
  Hao Wei$^{1}$,
  Peiji Wang$^{1}$,
  Qianhao Wang$^{1}$,
  Tong Qin$^{2}$,
  Fei Gao$^{1}$ and 
  Yulin Si$^{1}$
\thanks{$^1$Hao Wei, Peiji Wang, Qianhao Wang, Fei Gao, and Yulin Si are with Zhejiang University, Hangzhou, China. {\tt\small  \{isweihao, qianhaowang, fgaoaa, yulinsi\}@zju.edu.cn}}
\thanks{$^2$Tong Qin is with the Global Institute of Future Technology, Shanghai Jiao Tong University, Shanghai, China.
		{\tt\small  \{qintong\}@sjtu.edu.cn}. 
}
}

\begin{document}

\maketitle
\thispagestyle{empty}
\pagestyle{empty}

\begin{abstract}

Underwater environments impose severe challenges to visual–inertial odometry systems, as strong light attenuation, marine snow and turbidity, together with weakly exciting motions, degrade inertial observability and cause frequent tracking failures over long-term operation. While tightly coupled acoustic–visual–inertial fusion, typically implemented through an acoustic Doppler Velocity Log (DVL) integrated with visual–inertial measurements, can provide accurate state estimation, the associated graph-based optimization is often computationally prohibitive for real-time deployment on resource-constrained platforms. Here we present FAR-AVIO, a Schur-Complement based, tightly coupled acoustic-visual-inertial odometry framework tailored for underwater robots. FAR-AVIO embeds a Schur complement formulation into an Extended Kalman Filter(EKF), enabling joint pose–landmark optimization for accuracy while maintaining constant-time updates by efficiently marginalizing landmark states. On top of this backbone, we introduce Adaptive Weight Adjustment and Reliability Evaluation(AWARE), an online sensor health module that continuously assesses the reliability of visual, inertial and DVL measurements and adaptively regulates their sigma weights, and we develop an efficient online calibration scheme that jointly estimates DVL–IMU extrinsics, without dedicated calibration manoeuvres. Numerical simulations and real-world underwater experiments consistently show that FAR-AVIO outperforms state-of-the-art underwater SLAM baselines in both localization accuracy and computational efficiency, enabling robust operation on low-power embedded platforms. Our implementation has been released as open source software at \textcolor{red}{\href{https://far-vido.gitbook.io/far-vido-docs}{https://far-vido.gitbook.io/far-vido-docs}}

\end{abstract}

\section{INTRODUCTION}
\label{sec:introduction}

Marine robotics, such as Autonomous Underwater Vehicles (AUVs) and Remotely Operated Vehicles (ROVs), have become indispensable platforms for subsea infrastructure inspection, offshore energy maintenance, and ocean exploration\cite{aqualocdataset,TankDatasetUnderwaterxu2025a}. For these missions, accurate and drift-bounded localization is essential, yet satellite navigation is unavailable in underwater environments and external positioning infrastructure is typically absent or severely constrained in practice. Cameras, as compact and low-cost payloads, are therefore increasingly adopted on underwater robots, and visual Simultaneous Localization and Mapping (SLAM) has become a natural candidate for precise underwater localization.

State-of-the-art visual and visual–inertial SLAM systems, such as ORB-SLAM3\cite{campos2021orb}, VINS-Mono\cite{qin2018vins}, DM-VIO\cite{dmvio}, OV2-SLAM\cite{ov2}, have demonstrated impressive performance in ground and aerial scenarios with abundant visual texture and well-excited inertial motion. However, underwater environments present substantial challenges to these methods. Rapid light attenuation, poor illumination, and dense "marine snow" often cause visual degradation that persists for tens of seconds to minutes, exceeding the temporal horizon over which pure visual tracking remains reliable\cite{TankDatasetUnderwaterxu2025a}. Moreover, underwater robots frequently operate with gentle, quasi-static motions, leading to low accelerometer signal-to-noise ratios and difficult IMU initialization, as a result, conventional visual–inertial pipelines struggle to maintain consistent accuracy during prolonged visual degradation\cite{TightlyCoupledVisualInertialPressureFusionhu2022a,vip-init}.

\begin{figure}[t]
    \centering
    \includegraphics[width=1\linewidth]{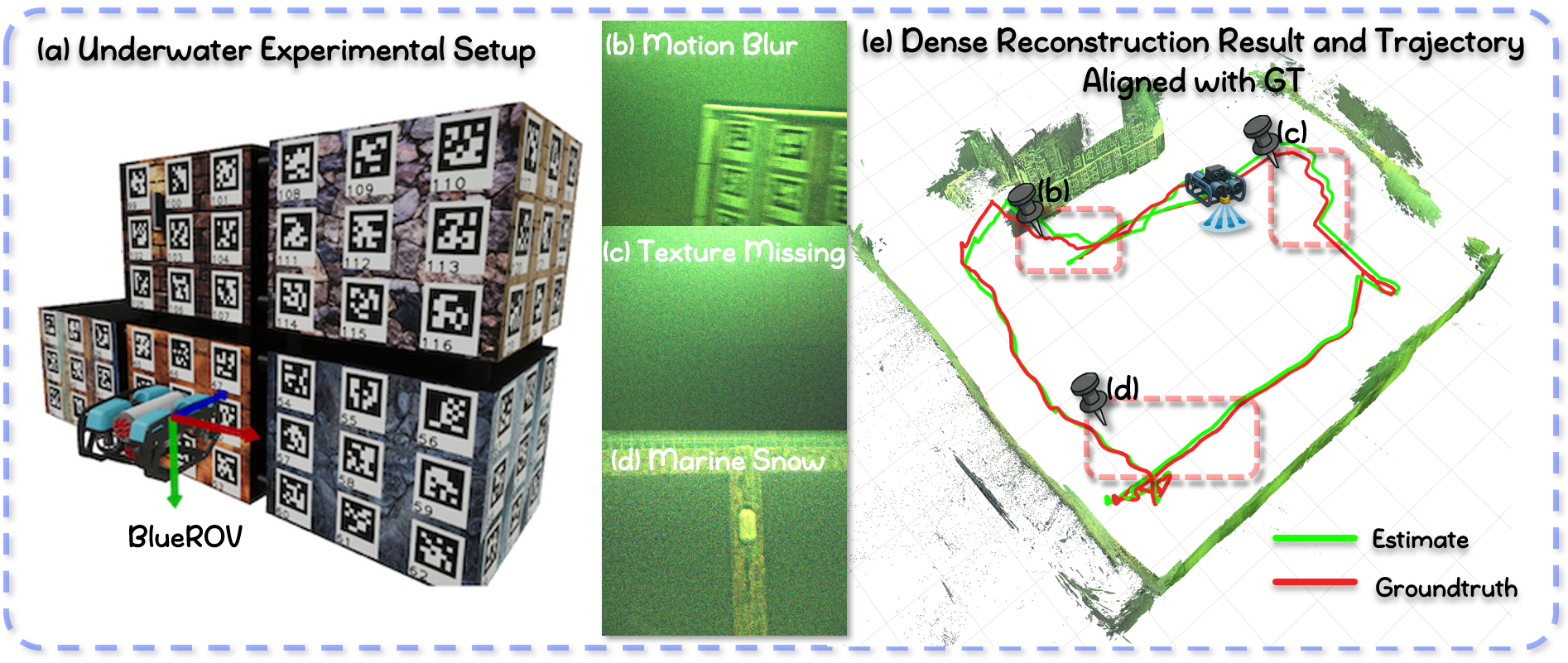}
    \caption{
        Real-world deployment of the proposed system in underwater environments. 
        (a) illustrates the experimental setup, where an ROV inspects the target underwater structure. 
        (b), (c), and (d) presents typical visual challenges encountered underwater, including motion blur, longtime textureless regions, and marine snow. 
        (e) shows the dense reconstruction result and estimated trajectory produced by FAR-AVIO.
    }
    \vspace{-4mm}
    \label{fig:run}
\end{figure}

To mitigate these limitations, numerous underwater SLAM and navigation frameworks have incorporated additional proprioceptive and exteroceptive sensors, such as DVL, depth/pressure sensors and sonar. Recent tightly coupled visual–inertial–DVL\cite{AQUASLAMTightlyCoupledxu2025} and sonar-based\cite{SVIn2MultisensorFusionbasedrahman2022a} systems have demonstrated remarkable robustness and accuracy in challenging underwater scenarios. However, most of these approaches rely on large-scale graph-based optimization or factor-graph SLAM backends, which incur substantial computational overhead and are difficult to deploy in real time on resource-constrained embedded platforms. In contrast, existing filter-based multi-sensor fusion methods typically simplify visual and DVL information into low-dimensional pose or velocity measurements\cite{TightlyCoupledVisualDVLzhao2023,WangsenUnderwaterVisualAcousticSLAM2021}, without fully exploiting point-level constraints or performing online calibration of multi-sensor extrinsics, which limits their ultimate accuracy and long-term consistency.

\begin{figure}[t]
    \centering
    \includegraphics[width=1\linewidth]{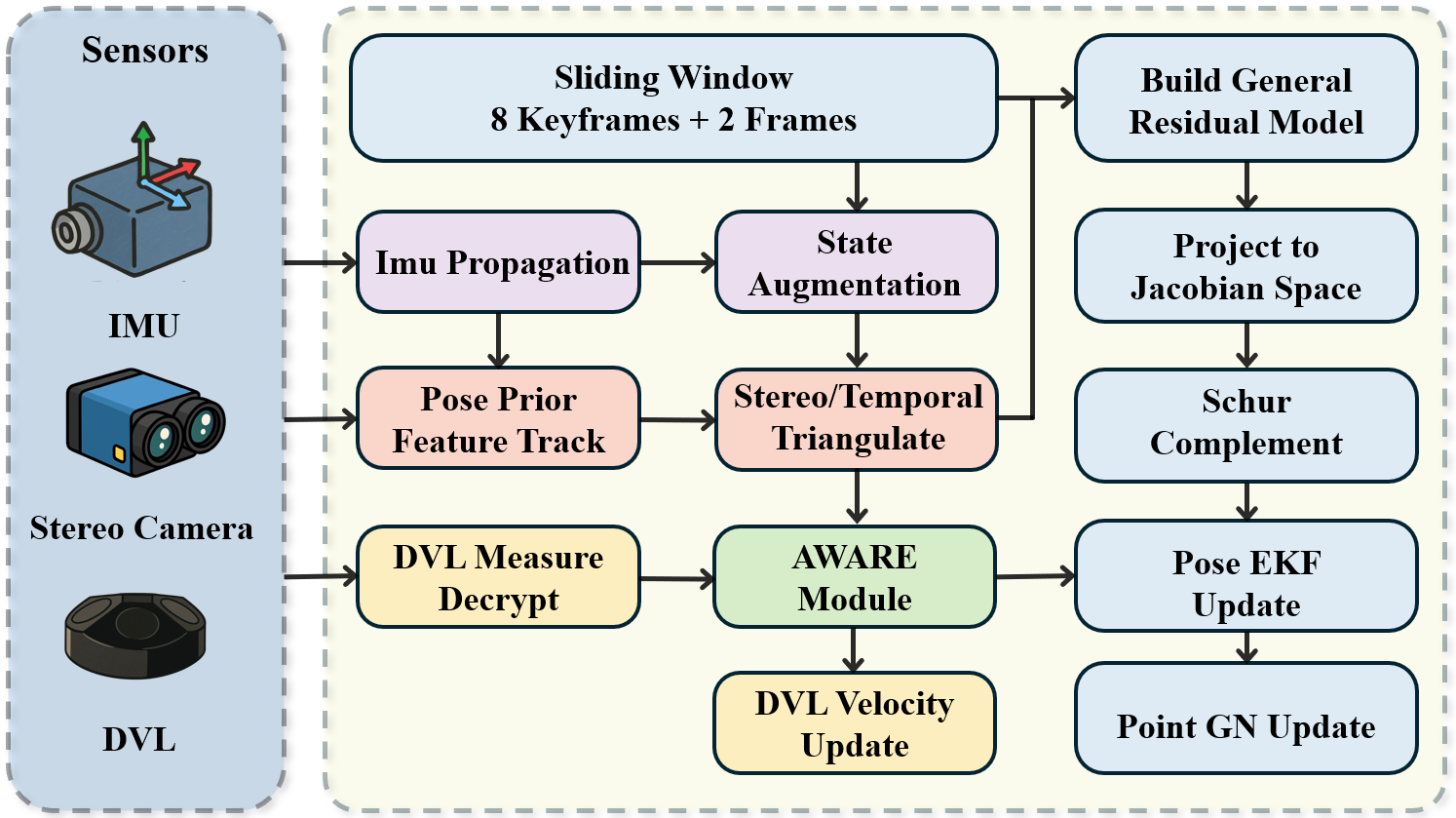}
    \caption{
        Proposed system architecture of FAR-AVIO. Input is from an underwater robot equipped with a stereo camera, an IMU, and a DVL.
    }\vspace{-5mm}
    \label{fig:system}
\end{figure}

These observations highlight a persistent gap between accuracy and efficiency in current underwater localization systems. To mitigate this gap, this paper presents FAR-AVIO, \textbf{a Fast and Robust Acoustic-Visual-Inertial Odometry with online calibration and an AWARE module}. FAR-AVIO is inspired by Schur-Complement based sliding-window filters \cite{SlidingWindowFiltersibley2010,fan2024schurvins}, and employs the Schur complement within an EKF framework, enabling joint pose–landmark optimization while leveraging landmark independence for efficient, constant-time updates. This design bridges the gap between optimization-based accuracy and filter-based efficiency. 
The main contributions of this work are as follows:
\begin{itemize}
  \item We propose a first Schur-Complement based tightly coupled Acoustic-Visual-Inertial odometry framework that rigorously models DVL measurements from doppler-shift principles and embeds them in filter based backend optimization, while jointly performing online DVL–IMU extrinsic calibration, thereby achieving high localization accuracy and real-time performance on resource-constrained underwater platforms without dedicated calibration procedures.

  \item FAR-AVIO introduces an AWARE module, performing an online health score mechanism that dynamically adjusts sensor fusion sigma scale based on real-time reliability assessment, enabling robust operation under sensor degradation and failure conditions.

  \item FAR-AVIO has been validated through extensive experiments including numerical simulations, and real-world underwater experiments, demonstrating superior accuracy and computational efficiency compared to state-of-the-art methods (see running result example in Fig.~\ref{fig:run}). To benefit the research community, we release the complete implementation as open-source software.
\end{itemize}
The proposed system architecture is illustrated in Fig.~\ref{fig:system}, and the rest of the paper is organized as follows: Section \ref{sec:related-work} reviews related literature. Section \ref{sec:methodology} describes the proposed FAR-AVIO framework, including the Schur-Complement based visual update, DVL measurement update, online sensor calibration, and the AWARE module. Section \ref{sec:experiments} presents experimental evaluation on real-world underwater datasets, Section \ref{sec:conclusion} concludes the paper and discusses future work.


\section{RELATED WORK}\label{sec:related-work}

We brieﬂy discuss the related works in this section. Visual and visual–inertial SLAM have been extensively studied in terrestrial and aerial domains\cite{campos2021orb,geneva2020openvins,qin2018vins,kumar-msckf,ov2,dmvio,svo,okvis2014}. These approaches provide a strong baseline when visual texture is abundant and inertial excitation is sufficient, but their performance degrades severely in underwater environments as discussed in the Section \ref{sec:introduction}. In the following, we focus on underwater multi-sensor fusion methods, with particular emphasis on DVL-based localization.

Early underwater SLAM frameworks combined a DVL, stereo vision and gyroscopic measurements in a loosely integrated pipeline\cite{WangsenUnderwaterVisualAcousticSLAM2021,xushida-underwater-visual-acoustic-slam-with-extrinsic-calibration-2021}, where DVL outputs served only as external velocity priors to the visual estimator. Although this configuration can deliver reasonable pose estimates in difficult underwater scenes, it neither exploits accelerometer data nor models gyroscope bias, which leads to drifting roll and pitch and ultimately limits the long-term consistency of the solution. To improve this, several tightly coupled visual–DVL and visual–inertial–DVL frameworks have been proposed. A visual–DVL fusion method in\cite{Visual-DVL-Fusion-HuangyuPei} directly injects DVL velocities into a factor-graph backend to jointly optimize camera poses and DVL measurements, but does not incorporate IMU data, restricting robustness under rapid motion and severe visual degradation. A tightly coupled visual–inertial–DVL odometry based on filter \cite{TightlyCoupledVisualDVLzhao2023} jointly fuses all three modalities in a filter-based backend, yet assumes known, fixed extrinsics, making it sensitive to calibration drift or hardware changes.

More recent work extends DVL fusion to richer multi-sensor SLAM systems. A graph-based LiDAR–VI–DVL framework for autonomous surface vehicles\cite{lidar-dvl-fusion-thoms} achieves accurate localization on water-surface trajectories, but is tailored to 2-D surface motion with strong LiDAR constraints and is not directly applicable to fully submerged 3-D trajectories. A tightly coupled visual–inertial–acoustic system\cite{HuangYupei-TC} and AQUA-SLAM\cite{AQUASLAMTightlyCoupledxu2025} further integrate DVL, camera and IMU in ORB-SLAM3-style backends with online extrinsic calibration, and the latter provides the Tank dataset\cite{TankDatasetUnderwaterxu2025a} for benchmarking. However, these factor-graph-based methods entail significant computational cost and are challenging to deploy on resource-constrained platforms for real-time applications.


\section{FILTER DESCRIPTION}\label{sec:methodology}

\subsection{State Definition and Propagation}

Following the state formulation in~\cite{eskf-solar} and augmenting it with sensor extrinsic parameters, we define the system state as
\begin{equation}
\boldsymbol{x}_{b} =
\left[
\boldsymbol{p}^w_b \ \;
\boldsymbol{v}^w_b \ \;
\boldsymbol{R}^w_b \ \;
\boldsymbol{b}_{a} \ \;
\boldsymbol{b}_{g} \ \;
\boldsymbol{T}^b_c \ \;
\boldsymbol{T}^b_D
\right]^{\top},
\end{equation}
where $\boldsymbol{p}^w_b$, $\boldsymbol{v}^w_b$, and $\boldsymbol{R}^w_b$ denote the position, velocity, and orientation of the body frame \{$b$\} expressed in the world frame \{$w$\}, respectively. The vectors $\boldsymbol{b}_{a}$ and $\boldsymbol{b}_{g}$ represent the accelerometer and gyroscope biases of the IMU. The terms $\boldsymbol{T}^b_c$ and $\boldsymbol{T}^b_D$ denote the extrinsic transformations from the body frame \{$b$\} to the camera frame \{$c$\} and the DVL frame \{$D$\}, respectively, each consisting of a rotation and translation (e.g., $\boldsymbol{T}^b_c = \{\boldsymbol{R}^b_c,\boldsymbol{p}^b_c\}$). Accordingly, the associated error-state vector $\delta \boldsymbol{x}_{b}$ is defined as
\begin{equation}
\delta \boldsymbol{x}_{b} =
\left[
\delta \boldsymbol{p}^w_b \ \;
\delta \boldsymbol{v}^w_b \ \;
\delta \boldsymbol{\theta}^w_b \ \;
\delta \boldsymbol{b}_{a} \ \;
\delta \boldsymbol{b}_{g} \ \;
\delta \boldsymbol{\mathcal{T}}^b_c \ \;
\delta \boldsymbol{\mathcal{T}}^b_D
\right]^{\top}.
\end{equation}
For non-rotational part, the standard additive error model $\boldsymbol{x} = \hat{\boldsymbol{x}} + \delta \boldsymbol{x}$ is adopted, where $\hat{\boldsymbol{x}}$ denotes the nominal state. For rotational components, perturbations are defined on $\mathrm{SO}(3)$ using the approximation
\begin{equation}
\boldsymbol{R} = \operatorname{Exp} \!\left(\left\lfloor \delta \boldsymbol{\theta} \right\rfloor_{\times}\right) \hat{\boldsymbol{R}},
\end{equation}
where $\delta \boldsymbol{\theta} \in \mathbb{R}^3$ is minimal rotation error representation, $\left\lfloor \cdot \right\rfloor_{\times}$ denotes skew-symmetric operator, and $\operatorname{Exp} (\cdot)$ is the matrix exponential operation. $\delta \boldsymbol{\mathcal{T}} \in \mathbb{R}^6$ is the minimal perturbations of the camera and DVL extrinsic transformations, each stacking a translation and a rotation error state.

The continuous-time dynamics of the nominal state are given by equation ~\eqref{con:nominal-dynamics}
\begin{equation}
\begin{aligned}
\dot{\hat{\boldsymbol{p}}}^w_b & =\hat{\boldsymbol{v}}^w_b,\ \dot{\hat{\boldsymbol{b}}}_{a} =\boldsymbol{0},\ \dot{\hat{\boldsymbol{b}}}_{g} =\boldsymbol{0} \\
\hat{\boldsymbol{\mathcal{T}}}^b_d & =\boldsymbol{0},\ \hat{\boldsymbol{\mathcal{T}}}^b_d= \boldsymbol{0} \\
\dot{\hat{\boldsymbol{v}}}^w_b & =\hat{\boldsymbol{R}}^w_b\left(\tilde{\boldsymbol{a}}-\hat{\boldsymbol{b}}_{a}\right)+\mathbf{g}^w \\
\dot{\hat{\boldsymbol{R}}}^w_b & =\hat{\boldsymbol{R}}^w_b
\left\lfloor 
\left(\tilde{\boldsymbol{\omega}}-\hat{\boldsymbol{b}}_{g}\right)\right\rfloor_{\times}, \\
\end{aligned}
\label{con:nominal-dynamics}
\end{equation}
where $\tilde{\boldsymbol{a}}$ and $\tilde{\boldsymbol{\omega}}$ are the raw accelerometer and gyroscope measurements, respectively, and $\mathbf{g}^w$ is the gravity vector in the world frame. The error-state propagation can be derived by linearizing the equation ~\eqref{con:nominal-dynamics}, yielding
\begin{equation}
\delta\dot{{\boldsymbol{x}}}_{b}=\boldsymbol{F} \delta {\boldsymbol{x}}_{b}+\boldsymbol{G} \boldsymbol{n}_{b},
\end{equation}
where $\boldsymbol{F}$ is the state transition Jacobian, $\boldsymbol{G}$ is the noise input matrix. $\boldsymbol{n}_{b} = [\boldsymbol{n}_{a}^{\top}\;\boldsymbol{n}_{aw}^{\top}\;\boldsymbol{n}_{g}^{\top}\;\boldsymbol{n}_{gw}^{\top}]^{\top}$ stacks the processes noise, $\boldsymbol{n}_{a}$ and  $\boldsymbol{n}_{g}$ denote the Gaussian white noise of the accelerometer and gyroscope measurements, respectively, while $\boldsymbol{n}_{aw}$ and $\boldsymbol{n}_{gw}$ model the random-walk processes driving the accelerometer and gyroscope biases.

\subsection{Visual Measurement and Equivalent Residual Model}
Assume that the $j$-th 3D landmark $\hat{\boldsymbol{\xi}}_j^w \in \mathbb{R}^3$, expressed in the world frame $w$, is observed by the camera at keyframe $i$. The projection model can be written as
\begin{equation}
\begin{aligned}
\hat{\boldsymbol{z}}_{ij}
&=
\pi\!\left(
\hat{\boldsymbol{x}}_b,
\hat{\boldsymbol{\xi}}_j^w
\right),
\end{aligned}
\end{equation}
where $\hat{\boldsymbol{z}}_{ij} \in \mathbb{R}^2$ is the predicted pixel coordinate of the landmark in the image plane, $\pi(\cdot)$ denotes the camera projection function that maps the 3D landmark in the world frame to 2D pixel coordinates based on the current state estimate $\hat{\boldsymbol{x}}_b$. Linearizing around the nominal state yields the reprojection residual
\begin{equation}
\boldsymbol{r}_{ij} = \boldsymbol{z}_{ij} -
\hat{\boldsymbol{z}}_{ij}
\simeq
\boldsymbol{H}_{x,ij}\,\delta\boldsymbol{\mathcal{X}}
+
\boldsymbol{H}_{f,ij}\,\delta\boldsymbol{\xi}^w_j
+
\boldsymbol{n}_{ij},
\label{con:reprojection-residual}
\end{equation}
where $\boldsymbol{r}_{ij}$ and $\boldsymbol{z}_{ij}$ are the reprojection residual and visual measurement from the frontend track result, respectively, $\boldsymbol{n}_{ij}$ is the measurement noise, $\boldsymbol{H}_{x,ij}$ and $\boldsymbol{H}_{f,ij}$ are the Jacobians of the projection function with respect to the system state and the landmark position, and $\delta\boldsymbol{\mathcal{X}}$ and $\delta\boldsymbol{\xi}^w_j$ are the state and landmark position perturbations, respectively. Their explicit forms are given by
\begin{equation}
\begin{aligned}
\boldsymbol{H}_{x,ij}&=\boldsymbol{J}_\pi
\left[
\begin{array}{cc}
\bigl(\hat{\boldsymbol{R}}^{b}_{c}\bigr)^\top
{\left\lfloor \hat{\boldsymbol{\xi}}^b_j\right\rfloor}_{\times}
{{\boldsymbol{R}}^w_{b_i}}^{\top} &
-{\boldsymbol{R}^w_{c_i}}^{\top}
\end{array}
\right], \\
\boldsymbol{H}_{f,ij}&=
\left[
\boldsymbol{J}_\pi 
{\boldsymbol{R}^w_{c_i}}^{\top}
\right],
\end{aligned}
\end{equation}
$\boldsymbol{J}_\pi$ denotes the Jacobian of the camera pixel projection function with respect to the 3D point in the camera frame, and its explicit form depends on the specific camera model. Stacking all observations across multiple keyframes in sliding window with equation~\eqref{con:reprojection-residual}, we obtain
\begin{equation}
{\boldsymbol{r}}=\boldsymbol{H}_x\,\delta\boldsymbol{\mathcal{X}}+
\boldsymbol{H}_f\,\delta\boldsymbol{\xi}+\boldsymbol{n},
\label{con:stacked-reprojection}
\end{equation}
where $\boldsymbol{r}$ and $[\boldsymbol{H}_x \ \boldsymbol{H}_f]$ are the stacked residuals and Jacobians, respectively. The stacked noise term is
$\boldsymbol{n}=[u,u,\cdots,u]^{\top}$ with covariance $\boldsymbol{\mathcal{R}}=\text{diag}(u^2,u^2,\cdots,u^2)$. Then, we directly project the measurement model of equation ~\eqref{con:stacked-reprojection} into the Jacobian space $\left[\boldsymbol{H}_x \ \boldsymbol{H}_f\right]^{\top}$ to construct an equivalent observation model as
\begin{equation}
\left[
  \begin{array}{c}
    {\boldsymbol{H}_x}^{\top} \\
    {\boldsymbol{H}_f}^{\top}
  \end{array} 
\right] 
\boldsymbol{r} = 
\begin{array}{c}
  \left[
  \begin{array}{c}
    {\boldsymbol{H}_x}^{\top} \\
    {\boldsymbol{H}_f}^{\top}
  \end{array} 
\right] 
\left[
  \begin{array}{cc}
    {\boldsymbol{H}_x} & {\boldsymbol{H}_f}
  \end{array} 
\right] 
\end{array}
\left[
  \begin{array}{c}
    \delta\boldsymbol{\mathcal{X}} \\
    \delta\boldsymbol{\xi}
  \end{array}
\right] + \boldsymbol{n}',
\label{con:schur-reprojection}
\end{equation}
$\boldsymbol{n}'$ is the equivalent observation noise, the corresponding covarianceis are given by 
\begin{equation}
\boldsymbol{\mathcal{R}}' = 
\left[
  \begin{array}{c}
    \boldsymbol{H}_x^{\top} \\
    \boldsymbol{H}_f^{\top}
  \end{array}
\right] 
\boldsymbol{\mathcal{R}}
\left[
  \begin{array}{cc}
    \boldsymbol{H}_x & \boldsymbol{H}_f
  \end{array}
\right].
\label{con:schur-noise-covariance}
\end{equation}
Equations~\eqref{con:schur-reprojection} and~\eqref{con:schur-noise-covariance} thus define an equivalent observation model and noise model in Jacobian space that fully preserve the information from all visual measurements, expanding getting rid of the original measurement dimension limitation. For clarity, expanding equation~\eqref{con:schur-reprojection} as
\begin{equation}
\begin{aligned}
\underbrace{\left[\begin{array}{c}
\boldsymbol{H}_{x}^{\top} \boldsymbol{r} \\
\boldsymbol{H}_{f}^{\top} \boldsymbol{r}
\end{array}\right]}_{\left[\begin{array}{c}
\boldsymbol{b}_{1} \\
\boldsymbol{b}_{2}
\end{array}\right]}=\underbrace{\left[\begin{array}{cc}
\boldsymbol{H}_{x}^{\top} \boldsymbol{H}_{x} & \boldsymbol{H}_{x}^{\top} \boldsymbol{H}_{f} \\
\boldsymbol{H}_{f}^{\top} \boldsymbol{H}_{x} & \boldsymbol{H}_{f}^{\top} \boldsymbol{H}_{f}
\end{array}\right]}_{\left[\begin{array}{cc}
\boldsymbol{C}_{1} & \boldsymbol{C}_{2} \\
\boldsymbol{C}_{2}^{\top} & \boldsymbol{C}_{3}
\end{array}\right]}\left[\begin{array}{c}
{\delta \boldsymbol{\mathcal{X}}} \\
{\delta \boldsymbol{\xi}}_{}
\end{array}\right]+\underbrace{\boldsymbol{n}^{\prime}}_{\left[\begin{array}{c}
\boldsymbol{n}_{1}^{\prime} \\
\boldsymbol{n}_{2}^{\prime}
\end{array}\right]}.
\end{aligned}
\label{equ:schur-reprojection-expanded}
\end{equation}

Due to the limitation of the dimension in the system, the filter state does not include the landmark perturbation $\delta\boldsymbol{\xi}$, a standard observation model must be constructed by marginalizing $\delta\boldsymbol{\xi}$. To this end, adopt a Schur-Complement based elimination~\cite{SlidingWindowFiltersibley2010} to eliminate the landmark state from the observation model in equation~\eqref{equ:schur-reprojection-expanded}, yielding
\begin{subequations}
\begin{align}
{\left[\boldsymbol{b}_{1}-\boldsymbol{C}_{2} \boldsymbol{C}_{3}^{-1} \boldsymbol{b}_{2}\right] } & =\left[\boldsymbol{C}_{1}-\boldsymbol{C}_{2} \boldsymbol{C}_{3}^{-1} \boldsymbol{C}_{2}^{\top}\right] \delta{\boldsymbol{\mathcal{X}}}+\boldsymbol{n}_{1}^{\prime \prime}, \label{equ:schur-eliminated-blocks-1}\\
\boldsymbol{\mathcal{R}}_{1}^{\prime \prime} & =\left[\boldsymbol{C}_{1}-\boldsymbol{C}_{2} \boldsymbol{C}_{3}^{-1} \boldsymbol{C}_{2}^{\top}\right] u^{2}, \label{equ:schur-eliminated-blocks-2} 
\end{align}
\label{equ:schur-eliminated-blocks}
\end{subequations}
where $\boldsymbol{n}_{1}^{\prime \prime}$ is the resulting equivalent observation noise with covariance $\boldsymbol{\mathcal{R}}_{1}^{\prime \prime}$. Equations~\eqref{equ:schur-eliminated-blocks-1} and~\eqref{equ:schur-eliminated-blocks-2} define an equivalent observation model and noise model that depend only on the error state $\delta{\boldsymbol{\mathcal{X}}}$, fully preserving the information from all visual measurements while successfully marginalizing out the landmark state $\delta{\boldsymbol{\xi}}$. The resulting equivalent residual can then be inserted into the standard EKF update procedure.

\subsection{DVL Measurement Model}

\subsubsection{Single-Beam Doppler Velocity}

\begin{figure}[tbp]
    \centering
    \includegraphics[width=0.99\linewidth]{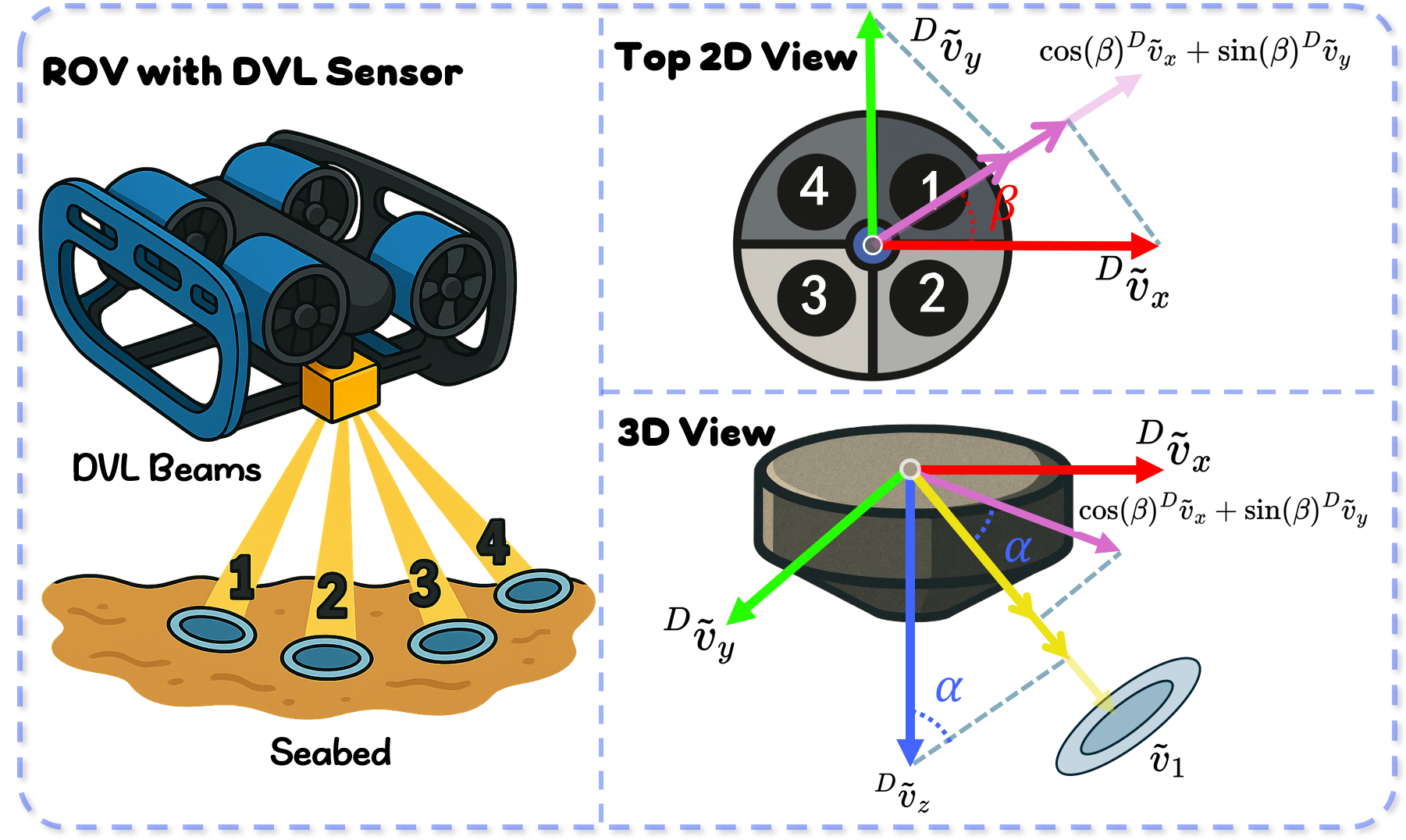}
    \caption{\label{fig:dvl-measure}DVL transducer measurements are illustrated in both 2-D and 3-D views. The instrument comprises four transducers oriented in different directions, with Transducer~1 shown as a representative example.}
\end{figure}

A DVL transducer emits a narrowband acoustic wave of known carrier frequency $f_t$ and receives the echo reflected by the seabed or water-borne scatterers. Let $f_r$ denote the received frequency and $\Delta f = f_r - f_t$ denote measured doppler shift. Under the standard narrowband, small-velocity assumption $\lvert v_r \rvert \ll c_s$, where $c_s$ denotes the speed of sound in water, the classic monostatic doppler relation yields the radial velocity $v_r$ along the acoustic beam as
\begin{equation}
  v_r \approx -\,\frac{c_s}{2 f_t}\,\Delta f.
  \label{eq:dvl-doppler}
\end{equation}
Here the sign convention is chosen such that $v_r > 0$ corresponds to the vehicle moving towards the seabed along the beam direction. For each of the four DVL beams (indexed by $i=1,\dots,4$, as shown in Fig.~\ref{fig:dvl-measure}), let $\tilde v_i$ denote the scalar radial velocity obtained by converting its measured doppler shift via equation \eqref{eq:dvl-doppler}. Ideally, $\tilde v_i$ coincides with the true radial velocity $v_{r,i}$ along that beam, in practice, it is corrupted by noise and occasional outliers due to low signal-to-noise ratio or loss of bottom lock. We model the single-beam measurement as
\begin{equation}
  \tilde v_i = v_{r,i} + n_i,\quad n_i \sim \mathcal{N}(0,\sigma_i^2),
  \label{eq:dvl-single-beam-scalar}
\end{equation}
where $n_i$ denotes zero-mean Gaussian measurement noise with variance $\sigma_i^2$. 

\subsubsection{DVL Velocity Measurement Model}
The linear velocity at the DVL acoustic center, expressed in the DVL frame $\{D\}$, as
\begin{equation}
  {}^{D}\boldsymbol{\tilde v}
  =
  \bigl[
    {}^{D}\tilde v_x,\,
    {}^{D}\tilde v_y,\,
    {}^{D}\tilde v_z
  \bigr]^\mathsf{T}.
\end{equation}
Following the DVL geometry in Fig.~\ref{fig:dvl-measure}, the orientation of each beam $\boldsymbol{e}_i$ respect to the frame $\{D\}$ has been parameterized by a fixed tilt angle $\alpha$ from the horizontal plane $x_Dy_D$ and an azimuth angle $\beta_i$ about the $z_D$-axis. The azimuth $\beta_i$ is measured in the $x_Dy_D$-plane from the $x_D$-axis to the projection of the $i$-th beam. For a four-beam Janus configuration, the beams share the same tilt $\alpha$ and have azimuths $\beta_i$. The example $\boldsymbol{e}_i$ of transducer $1$ can be expressed as
\begin{equation}
  \boldsymbol{e}_1
  =
  \bigl[
    \cos\beta_1 \cos\alpha,\,
    \sin\beta_1 \cos\alpha,\,
    \sin\alpha
  \bigr]^\mathsf{T}.
\end{equation}
Then the true radial velocity measured by this beam is simply the projection of the 3D velocity onto the beam direction:
\begin{equation}
  v_{r,i} = \boldsymbol{e}_i^\top \, {}^{D}\tilde{\boldsymbol{v}}.
  \label{eq:dvl-single-beam-proj}
\end{equation}
Substituting \eqref{eq:dvl-single-beam-proj} into \eqref{eq:dvl-single-beam-scalar}, the single-beam DVL measurement model can be written as
\begin{equation}
  \tilde v_i = \boldsymbol{e}_i^\top \, {}^{D}\tilde{\boldsymbol{v}} + n_i,
  \label{eq:dvl-single-beam-model}
\end{equation}
which states that each transducer provides a observation of the vehicle velocity component along its own acoustic axis.

Stacking the four scalar beam measurements into a vector ${}^{b}\tilde{\boldsymbol{v}}
  =
  \bigl[
    \tilde v_1,\,
    \tilde v_2,\,
    \tilde v_3,\,
    \tilde v_4
  \bigr]^{\top}$ and collecting the beam direction vectors into a matrix, the single-beam model \eqref{eq:dvl-single-beam-model} can be written in compact form as
\begin{equation}
  {}^{b}\tilde{\boldsymbol{v}}
  =
  \boldsymbol{E}\,{}^{D}\boldsymbol{\tilde v}
  +
  \boldsymbol{n}_{b},
  \label{eq:dvl-multi-beam}
\end{equation}
where
\(
  \boldsymbol{E}
  =
  \bigl[
    \boldsymbol{e}_1,\,
    \boldsymbol{e}_2,\,
    \boldsymbol{e}_3,\,
    \boldsymbol{e}_4
  \bigr]^\top
  \in \mathbb{R}^{4\times 3}
\)
is the beam direction matrix, and
\(
  \boldsymbol{n}_{b}
  =
  \bigl[
    n_1,\,
    n_2,\,
    n_3,\,
    n_4
  \bigr]^\top
\)
is the stacked noise vector. The matrix $\boldsymbol{E}$ is full column rank (rank~3) as long as the beams are not coplanar, which holds for standard DVL configurations. Therefore, the 3D velocity ${}^{D}\boldsymbol{\tilde v}$ can be uniquely determined from the four beam measurements by solving the overdetermined linear system \eqref{eq:dvl-multi-beam} in a least-squares sense:
\begin{equation}
  {}^{D}\boldsymbol{\tilde v}
  =
  \bigl(\boldsymbol{E}^\top \boldsymbol{E}\bigr)^{-1}
  \boldsymbol{E}^\top
  {}^{b}\tilde{\boldsymbol{v}},
  \label{eq:dvl-velocity-least-squares}
\end{equation}
where the matrix inverse $\bigl(\boldsymbol{E}^\top \boldsymbol{E}\bigr)^{-1}$ depends only on the known beam geometry and can be pre-computed offline.

Since we are assuming independent Gaussian noise on each beam in equation \eqref{eq:dvl-single-beam-scalar}, we have
\begin{equation}
  \boldsymbol{n}_{b}
  \sim
  \mathcal{N}\bigl(\boldsymbol{0},\boldsymbol{\Sigma}_{b}\bigr),
  \quad
  \boldsymbol{\Sigma}_{b}
  =
  \mathrm{diag}(\sigma_1^2,\sigma_2^2,\sigma_3^2,\sigma_4^2).
\end{equation}
Substituting \eqref{eq:dvl-multi-beam} into \eqref{eq:dvl-velocity-least-squares} and separating the true velocity and noise terms yields
\begin{equation}
  {}^{D}\boldsymbol{\tilde v}
  =
  {}^{D}\boldsymbol{v}_{\text{true}}
  +
  \boldsymbol{A}\,\boldsymbol{n}_{b},
  \quad
  \boldsymbol{A}
  =
  \bigl(\boldsymbol{E}^\top \boldsymbol{E}\bigr)^{-1}
  \boldsymbol{E}^\top,
\end{equation}
where ${}^{D}\boldsymbol{v}_{\text{true}}$ denotes the true DVL-frame velocity. Since $\boldsymbol{n}_{b}$ is zero-mean Gaussian, its linear image $\boldsymbol{A}\boldsymbol{n}_{b}$ is also zero-mean Gaussian. Consequently, the estimated DVL-frame velocity remains Gaussian:
\begin{equation}
  {}^{D}\boldsymbol{\tilde v}
  \sim
  \mathcal{N}\bigl(
    {}^{D}\boldsymbol{v}_{\text{true}},\,
    \boldsymbol{\Sigma}_{D}
  \bigr),
  \quad
  \boldsymbol{\Sigma}_{D}
  =
  \boldsymbol{A}\,\boldsymbol{\Sigma}_{b}\,\boldsymbol{A}^\top.
\end{equation}
In the common case of identical per-beam variance $\sigma_i^2 = \sigma^2$, this simplifies to
\begin{equation}
  \boldsymbol{\Sigma}_{D}
  =
  \sigma^2
  \bigl(\boldsymbol{E}^\top \boldsymbol{E}\bigr)^{-1},
\end{equation}
which provides a convenient closed-form expression for the covariance of the DVL-frame velocity measurement.

\subsubsection{DVL Residual for ESKF State Update}
Given the estimated state, the predicted DVL-frame velocity is computed as
\begin{equation}
{}^{D}\hat{\boldsymbol{v}}
=
\left(\hat{\boldsymbol{R}}^{b}_{D}\right)^\top
\left(
\left(\hat{\boldsymbol{R}}^{w}_{b}\right)^\top \hat{\boldsymbol{v}}^{w}_{b}
+
\lfloor {}^{b}\hat{\boldsymbol{\omega}} \rfloor_\times \hat{\boldsymbol{p}}^{b}_{D}
\right),
\label{eq:dvl-predicted-velocity}
\end{equation}
where ${}^{b}\hat{\boldsymbol{\omega}}$ is the bias-corrected angular velocity in the body frame, $\hat{\boldsymbol{p}}^{b}_{D}$ is the IMU--DVL lever arm expressed in the body frame, and $\hat{\boldsymbol{R}}^{b}_{D}$ is the rotation from DVL to body frame. Thus we can get the DVL residual for the state update, according to the equation \eqref{eq:dvl-predicted-velocity} and \eqref{eq:dvl-velocity-least-squares}
\begin{equation}
  \boldsymbol{r}_{\mathrm{DVL}}
  =
  {}^{D}\tilde{\boldsymbol{v}}
  -
  {}^{D}\hat{\boldsymbol{v}}.
\end{equation}
This residual is then linearized with respect to the error state and used in the standard ESKF update with measurement covariance $\boldsymbol{\Sigma}_D$.


\subsection{AWARE: Adaptive Weight Adjustment and Reliability Evaluation}
Most fusion-based filters assume fixed, time-invariant measurement noise for each sensor, which is rarely valid in practice: visual quality changes with texture, illumination and motion, and DVL measurements degrade under poor bottom lock, scattering or flow disturbances. If such variations are ignored, bursts of bad measurements from one modality can corrupt the entire estimate even when other sensors remain reliable. AWARE addresses this by continuously assessing visual and DVL quality, adapting their effective covariances, and temporarily disabling severely degraded sensors, thus preventing any single faulty source from dominating the fusion.

For each sensor $s \in {\text{VIS},\text{DVL}}$, AWARE maintains a reliability scale $\sigma_s$ and a fixed-length queue $Q_s$ of recent ``unhealthy'' events. At every measurement, a sensor-specific quality score $q_s \in [0,1]$ is computed ($q_{\text{VIS}}$ from feature tracking statistics and reprojection error, $q_{\text{DVL}}$ from velocity consistency and DVL residuals). These scores drive covariance scaling and sensor gating decisions, as summarized in Algorithm~\ref{alg:aware}.
\begin{algorithm}[t]
  \caption{AWARE update for a generic sensor $s$}
  \label{alg:aware}
  \DontPrintSemicolon

  \KwIn{
    sensor stream $\{(t,\boldsymbol{z}_s)\}$,
    unhealthy events $Q_s$.\quad
  }
  \KwOut{
    adapted scale $\sigma_s$.\quad
  }

  \ForEach{measurement $(t,\boldsymbol{z}_s)$}{
    $q_s \gets \text{QualityScore}_s(\boldsymbol{z}_s)$\;

    \If{$enabled_s$}{
      \If{$q_s \ge \tau_s$}{
        \tcp{healthy}
        $\boldsymbol{R}_s^{\text{eff}} \gets \boldsymbol{R}_s / \sigma_s^2$\;
        \textsc{ESKF\_UPDATE}$(\boldsymbol{z}_s,\boldsymbol{R}_s^{\text{eff}})$\;
      }
      \Else{
        \tcp{unhealthy}
        append $(t,q_s)$ to $Q_s$\;
        $\sigma_s \gets \gamma_s \sigma_s$\;
        \If{$|Q_s| > N_s$}{
          remove oldest element from $Q_s$\;
        }
        \If{$|Q_s| = N_s$ \textbf{and} span$(Q_s) < \Delta T_s$}{
          $enabled_s \gets \text{false}$\;
          $\sigma_s \gets 1$, \quad $Q_s \gets \emptyset$\;
        }
        \textsc{ESKF\_UPDATE}$(\boldsymbol{z}_s,\sigma_s\boldsymbol{R}_s^{})$\;
      }
    }
    \Else{
      \tcp{sensor disabled}
      \If{$q_s \ge \tau_s^{\text{rec}}$}{
        $enabled_s \gets \text{true}$\;
        $\sigma_s \gets 1$, \quad $Q_s \gets \emptyset$\;
      }
    }
  }
\end{algorithm}

\subsection{Pose Prior Visual Frontend Tracking}

Our visual frontend is built on sparse Shi--Tomasi\cite{ShiTomasi} corners and an
optical–flow tracker augmented with an IMU-based pose prior. At each
new frame, candidate features are detected using the Shi--Tomasi
response on an image pyramid and distributed across the image to ensure
sufficient coverage. For frame-to-frame tracking, we adopt pyramidal
Lucas–Kanade\cite{Lucas-optical-flow} optical flow, but initialize the search by projecting the
3D landmarks from the previous frame to the current image using the
IMU-predicted relative pose. This pose prior shrinks the search region
around the expected pixel location and significantly improves tracking
robustness under fast motion and motion blur.

For stereo observations with known extrinsics, we further constrain
the correspondence search along the epipolar line in the right image.
Instead of running unrestricted 2D optical flow between the stereo pair,
candidate matches are only evaluated within a small window on the
epipolar line induced by the left-image feature. In practice we observe
that this epipolar-guided matching is crucial for long-baseline stereo
setups, where pure optical flow often fails due to large disparities and
strong perspective changes.


\section{Experiments}\label{sec:experiments}

\subsection{Experiment Settings and Datasets}

In this section, we evaluate the proposed FAR-AVIO on the public Tank dataset\cite{TankDatasetUnderwaterxu2025a}, which provides synchronized stereo, IMU, DVL and depth measurements collected in a wave tank. Accurate ground-truth (GT) camera poses are generated by the TankGT pipeline using AprilTag markers mounted on the underwater structure, enabling quantitative benchmarking under realistic underwater conditions. The eight sequences are grouped into three trajectory types (Structure, HalfTank, and WholeTank) with varying difficulty levels (Easy/Medium/Hard) determined by vehicle speed, lighting and the amount of textureless area. Fig.~\ref{fig:run} illustrates typical visual challenges and running result in the {HalfTank--Easy} sequence.

\subsection{Localization Performance Comparison}

\begin{table*}[t]
  \centering
  \caption{AVERAGE TRANSLATIONAL RMSE/STD (m) ON TANK SEQUENCES}
  \label{tab:uw-rmse}
  \begin{tabular}{lcccccc}
    \toprule
    Sequence & FAR-AVIO & AQUA-SLAM & UVA-SLAM & SVIN2  & ORB-SLAM3 & VINS-Fusion \\
    \midrule
    Structure Easy     & 0.11 / 0.03         & \textbf{0.07 / 0.03}   & 0.21 / 0.12  & 0.09 / 0.03   & 0.28 / 0.09  & 0.13  / 0.04  \\
    Structure Medium   & \textbf{0.16 / 0.05}         & 0.18 / 0.08   & 0.54 / 0.30  & 2.94 / 1.64   & 3.30 / 1.08  & NaN   / NaN   \\
    Structure Hard     & \textbf{0.13 / 0.04}         & 0.50 / 0.24   & 0.50 / 0.23  & 3.26 / 1.43   & 2.73 / 1.45  & NaN   / NaN   \\
    HalfTank Easy      & \textbf{0.19 / 0.10 }        & 0.28 / 0.17   & 1.69 / 1.10  & 6.01 / 4.33   & 2.69 / 1.45  & 1.24 / 0.30 \\
    HalfTank Medium    & 0.33 / 0.11         & \textbf{0.29 / 0.14}   & 0.44 / 0.22  & 3.40 / 1.76   & 0.74 / 0.38  & NaN   / NaN   \\
    HalfTank Hard      & \textbf{0.25 / 0.10}         & 0.36 / 0.22   & 0.58 / 0.37  & 77.6 / 55.07  & 1.10 / 0.70  & NaN   / NaN   \\
    WholeTank Medium   & \textbf{0.34 / 0.15}         & 0.52 / 0.28   & 1.34 / 0.73  & 0.72 / 0.41   & 1.18 / 0.71  & 13.08 / 7.22  \\
    WholeTank Hard     & 0.57 / 0.27         & \textbf{0.22 / 0.12}   & 1.11 / 0.83  & 0.83 / 0.65   & 2.96 / 2.49  & NaN   / NaN   \\
    \midrule
    Total RMSE           & \textbf{2.08} & 2.42        & 6.41  & 94.85  & 14.98 & NaN    \\
    \bottomrule
  \end{tabular}
\end{table*}

We have benchmark FAR-AVIO against five representative baselines: AQUA-SLAM, UVA-SLAM, SVIN2, ORB-SLAM3, and VINS-Fusion. For fairness, all methods use the same camera/IMU intrinsic and extrinsic parameters, and are run in stereo–inertial (or stereo–inertial–DVL when available) configurations. Estimated trajectories are aligned with ground truth using the method described in \cite{Umeyama}, the average root-mean-square error (RMSE) and standard deviation (STD) of the absolute translation error(ATE), summarized in Table~\ref{tab:uw-rmse}. Entries marked as NaN indicate repeated tracking failure or divergence before sequence completion.
\begin{figure*}[t]
  \centering
  \begin{subfigure}[t]{0.24\linewidth}
    \centering
    \includegraphics[width=\linewidth]{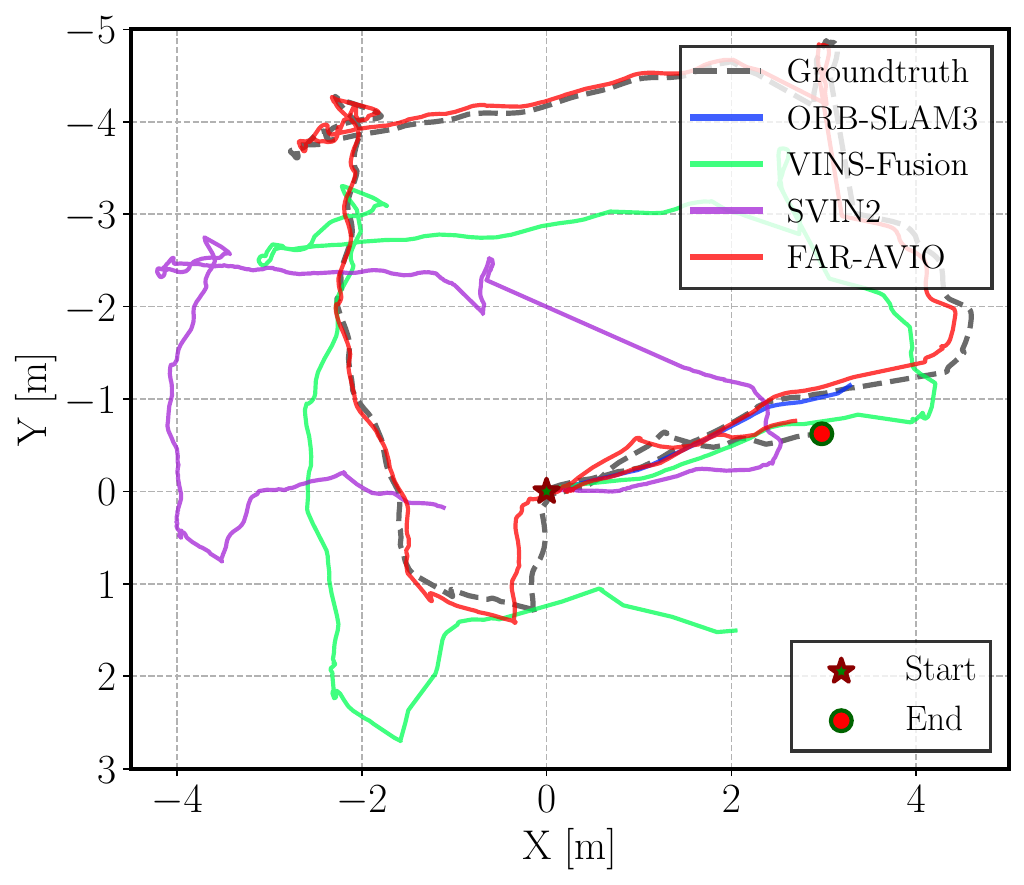}
    \subcaption{HalfTank--Easy}
    \label{fig:traj-compare-he}
  \end{subfigure}
  \hfill
  \begin{subfigure}[t]{0.24\linewidth}
    \centering
    \includegraphics[width=\linewidth]{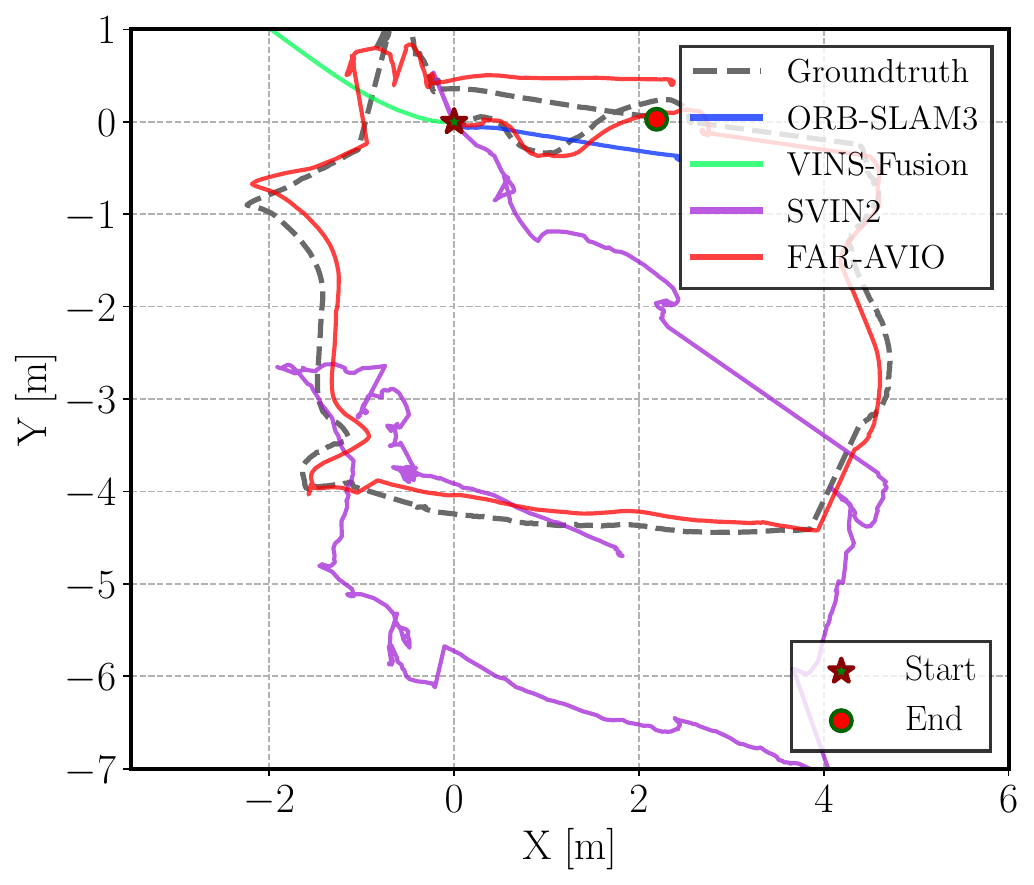}
    \subcaption{HalfTank--Medium}
    \label{fig:traj-compare-hm}
  \end{subfigure}
  \hfill
  \begin{subfigure}[t]{0.24\linewidth}
    \centering
    \includegraphics[width=\linewidth]{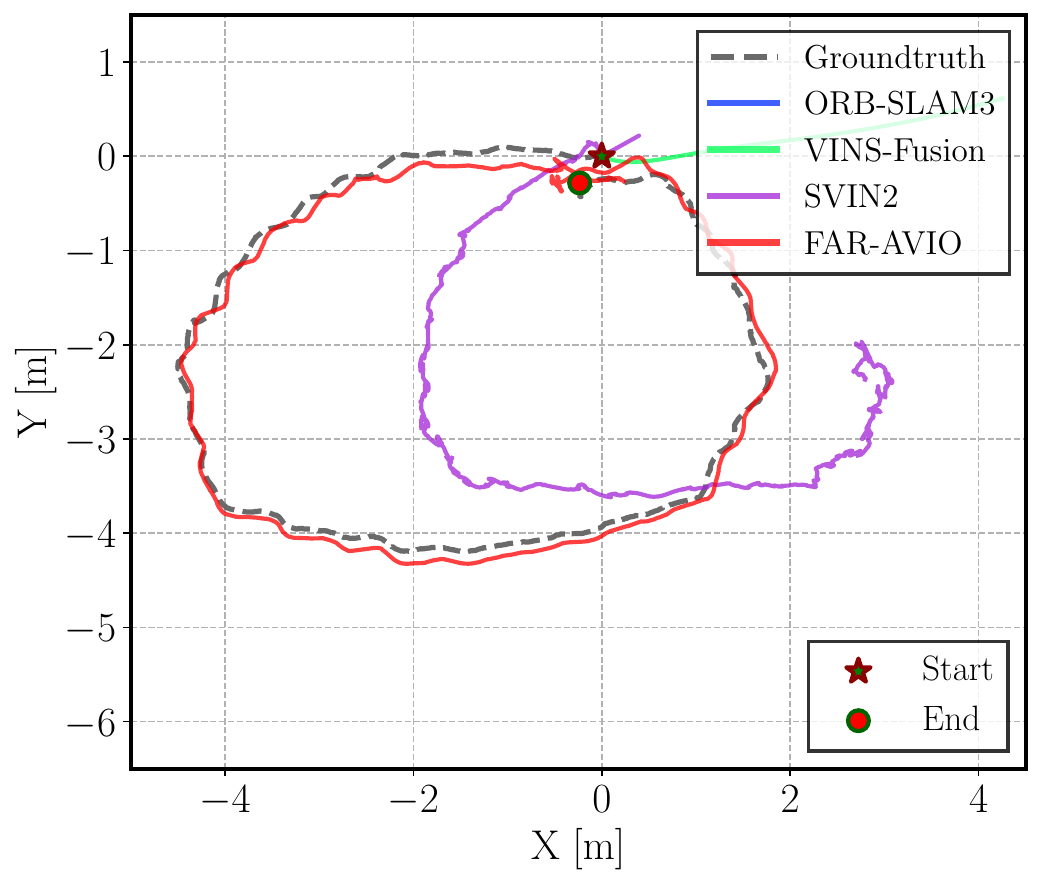}
    \subcaption{Structure--Medium}
    \label{fig:traj-compare-sm}
  \end{subfigure}
  \hfill
  \begin{subfigure}[t]{0.24\linewidth}
    \centering
    \includegraphics[width=\linewidth]{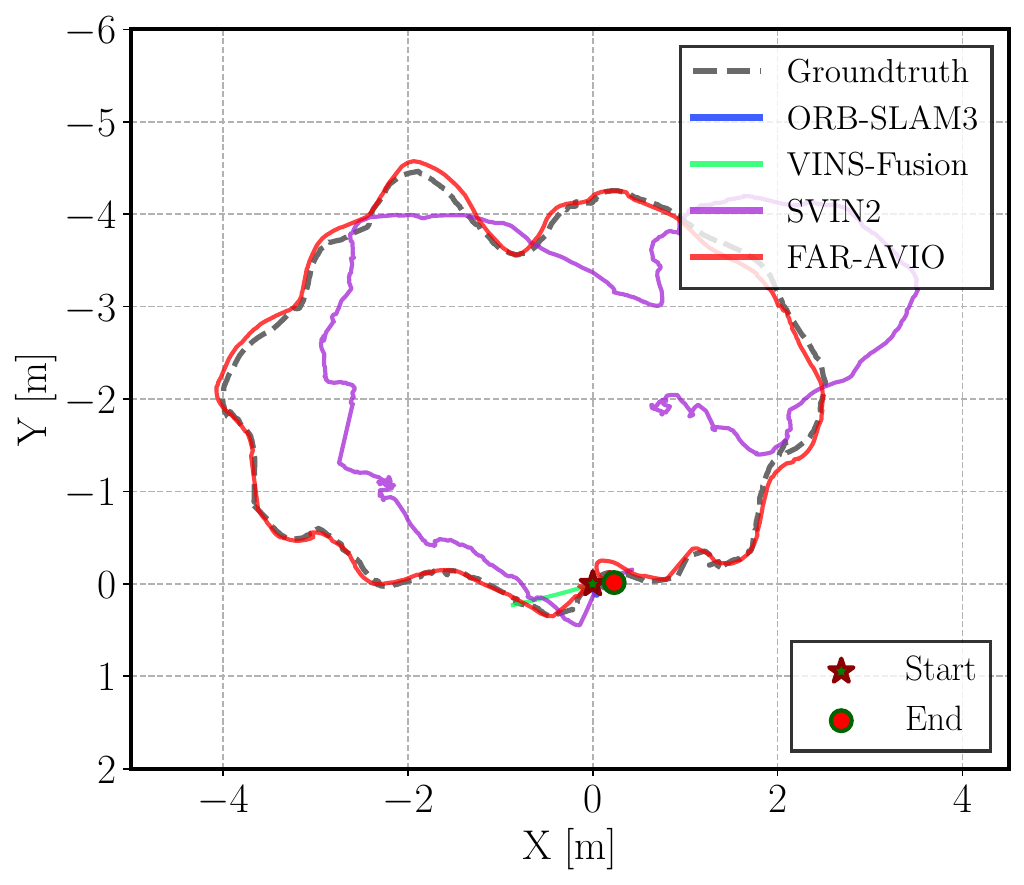}
    \subcaption{Structure--Hard}
    \label{fig:traj-compare-sh}
  \end{subfigure}
  \caption{Different baseline methods estimated trajectory comparisons.}
  \label{fig:traj-compare}
\end{figure*}

Overall, FAR-AVIO achieves the best average translational accuracy among all methods that complete all sequences, and consistently ranks first or second on every individual sequence. On the more challenging {Structure--Hard}, {HalfTank--Hard}, and {WholeTank--Medium} sequences, FAR-AVIO clearly outperforms the baselines, reducing the translational RMSE by up to about 75\% compared with AQUA-SLAM (e.g., $0.13$\,m vs.\ $0.50$\,m on {Structure--Hard}) and by more than an order of magnitude compared with purely visual–inertial methods. On the easier and medium sequences, AQUA-SLAM attains slightly lower RMSE on several cases, but FAR-AVIO remains competitive, with errors typically within a few centimeters of AQUA-SLAM while maintaining comparable or smaller standard deviations. This indicates that the proposed VI–DVL fusion system can match the accuracy of a heavy graph-optimization backend, while operating in a lightweight EKF framework. Figure~\ref{fig:traj-compare} visualizes representative trajectory comparisons on example sequences, for methods that diverged or failed to complete the sequence, only the successfully tracked portion of the trajectory is shown.

In contrast, ORB-SLAM3, VINS-Fusion, and SVIN2, which can only exploit stereo inertial data (no sonar, no pressure), frequently incur large drift or outright tracking failure in sequences with strong turbidity or prolonged visual degradation, as reflected by the meter-level errors and NaN entries (e.g., $29.83$\,m for VINS-Fusion on {HalfTank--Easy} and $77.6$\,m for SVIN2 on {HalfTank--Hard}). DVL-aided baselines UVA-SLAM and AQUA-SLAM substantially reduce drift compared with these pure VI methods and achieve sub-meter accuracy on most sequences, yet still exhibit noticeably higher errors or less stable performance than FAR-AVIO on the more difficult Medium and Hard settings.\vspace{-1mm}

\subsection{Runtime and Computational Load}

\begin{figure}[t]
    \centering
    \includegraphics[width=1\linewidth]{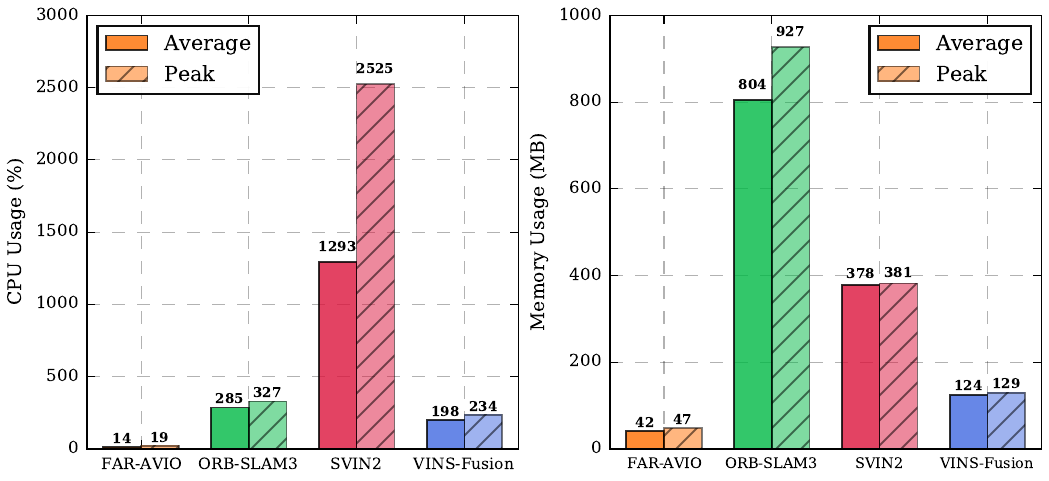}
    \caption{
        CPU load and memory usage of FAR-AVIO and baselines on Structure--Easy sequence.
    }
    \label{fig:cpu-load}
\end{figure}

We evaluate runtime and computational load for methods with public implementations (ORB-SLAM3, VINS-Fusion, SVIN2, FAR-AVIO) on a desktop CPU (AMD Ryzen 9 7950X, 32GB RAM) and an embedded platform (NVIDIA Jetson Orin NX, 8GB RAM). AQUA-SLAM (and UVA-SLAM), which are built on the ORB-SLAM-based backend, are not included in this comparison, but their computational cost is expected to be comparable to or higher than ORB-SLAM3. Figure~\ref{fig:cpu-load} shows that FAR-AVIO consistently exhibits the lowest CPU utilization and memory footprint among all baselines.

On the Orin NX, the per-module breakdown in Table~\ref{tab:runtime-breakdown-compare} shows that FAR-AVIO processes frames in $28.28$\,ms ($\approx 35$\,Hz), achieving about a $2.2\times$ speed-up over VINS-Fusion ($61.65$\,ms, $\approx 16$\,Hz). The main gain comes from the backend: VINS-Fusion spends $33.76$\,ms (54\%) in visual optimization, whereas FAR-AVIO visual update requires only $6.08$\,ms (21\%), and the additional DVL update adds merely $0.78$\,ms (2\%). As a result, the visual frontend becomes the dominant cost in FAR-AVIO, confirming that the proposed filter based backend effectively removes the optimization bottleneck and is well suited for embedded deployment.

\begin{table}[tbp]
  \centering
  \caption{Runtime comparison between VINS-Fusion and FAR-AVIO on the embedded platform.}
  \label{tab:runtime-breakdown-compare}
  \begin{tabular}{lcccc}
    \toprule
    \multirow{2}{*}{Module} & \multicolumn{2}{c}{VINS-Fusion} & 
    \multicolumn{2}{c}{FAR-AVIO (Ours)} \\
    \cmidrule(r){2-3} \cmidrule(l){4-5}
      & Time (ms) & Share & Time (ms) & Share \\
    \midrule
    Visual frontend        & 27.69     & 45\%    &  20.21 & 73\% \\
    Visual measure update  & 33.76     & 54\%    &  6.08  & 21\% \\
    DVL measure update     & --        & --      &  0.78  & 2\% \\
    Others (I/O, etc.)     & 0.20      & 1\%     &  1.21  & 4\% \\
    \midrule
    Total                  & 61.65     & 100\%    & 28.28 & 100\% \\
    \bottomrule
  \end{tabular}
\end{table}

\begin{figure}[!t]
  \centering
  \begin{subfigure}[t]{\linewidth}
    \centering
  \includegraphics[width=\linewidth]{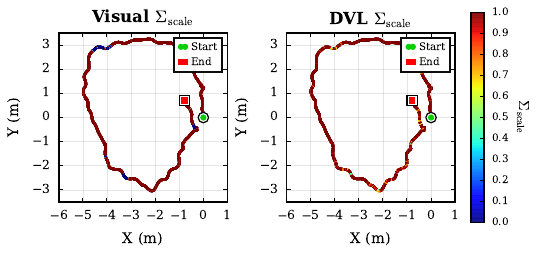}
    \subcaption{Structure--Easy Sequence}
    \label{fig:traj-sigma-se}
  \end{subfigure}
  \\
  \begin{subfigure}[t]{\linewidth}
    \centering
  \includegraphics[width=\linewidth]{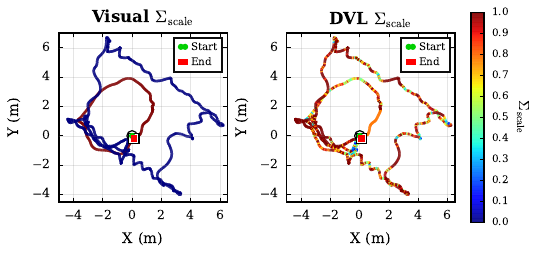}
    \subcaption{WholeTank--Hard Sequence}
    \label{fig:traj-sigma-wh}
  \end{subfigure}
  \caption{AWARE-estimated sigma scaling factors for visual and DVL measurements on two Tank sequences.}\vspace{-4mm}
  \label{fig:sigma-scalar}
\end{figure}

\subsection{Ablation Study for AWARE Modular and Extrinsic Calibration}

We conduct ablation experiments to quantify the contribution of the
proposed AWARE module and the online calibration of the IMU--DVL
extrinsics. The study includes both real Tank sequences and purely
numerical simulations with synthetic IMU\cite{geneva2020openvins} and DVL measurements, where
ground-truth extrinsics are known.

\paragraph{Effect of the AWARE module.}
To assess whether AWARE behaves as intended, we visualize the instantaneous
confidence scales applied to visual and DVL measurements along the estimated
trajectory. For two representative sequences, {Structure--Easy} (SE) and
{WholeTank--Hard} (WH), we plot the trajectory and color each point by the
corresponding visual and DVL $\Sigma_{\text{scale}} \in [0,1]$, where values
close to $1$ indicate high confidence (nominal weighting) and values approaching
$0$ indicate strong down-weighting of that sensor, see Fig.~\ref{fig:traj-sigma-se}
and Fig.~\ref{fig:traj-sigma-wh}, on the SE sequence, which features relatively clean water and stable lighting,
the visual frontend rarely experiences long-term degradation. Accordingly, both
the visual and DVL scales remain close to $1$ along almost the entire trajectory,
with only minor fluctuations (Fig.~\ref{fig:traj-sigma-se}). This indicates that
AWARE does not introduce unnecessary re-weighting when all sensors operate
nominally, and effectively reduces to a standard tightly coupled VI–DVL fusion
scheme on easy sequences.

In contrast, the WH sequence is considerably more challenging, strong turbidity,
non-uniform illumination, and large portions of the trajectory with weak or
missing image features lead to extended periods of unreliable visual tracking.
In these segments, the visual $\Sigma_{\text{scale}}$ is significantly reduced
along the corresponding parts of the trajectory, while the DVL scale remains
close to $1$ (Fig.~\ref{fig:traj-sigma-wh}). This behavior shows that AWARE
automatically down-weights visual updates when the frontend reports poor
tracking quality, and simultaneously leans more on the DVL constraints to
stabilize the state estimate. As visual conditions recover, the visual scale
smoothly returns towards $1$, and the fusion reverts to a more balanced
VI–DVL weighting. Overall, these qualitative results confirm that AWARE
adapts the contribution of visual and DVL measurements in a manner that is
consistent with the underlying sensing conditions, rather than relying on
fixed, hand-tuned sensor weights.

To quantitatively assess the impact of AWARE on localization performance, we compare FAR-AVIO with and without AWARE on the Tank sequences, as well as FAR-AIO, a variant that fuses only IMU and DVL (no visual measurements) with AWARE enabled. As summarized in Table~\ref{tab:ablation-aware}, FAR-AVIO with AWARE achieves the lowest RMSE on all sequences and is the only variant that successfully completes every run, while both FAR-AIO and FAR-AVIO without AWARE experience failures on the more challenging cases.

\begin{figure}[!t]
    \centering
    \includegraphics[width=1\linewidth]{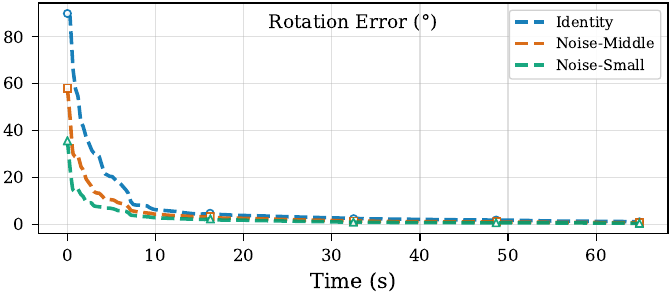}
    \caption{
        Extrinsic of rotation part calibration convergence from different initializations in numerical simulation.
    }
    \label{fig:dvl-extrinsic-calib}
\end{figure}

\begin{table}[t]
  \centering
  \caption{Ablation of the AWARE module on Tank sequences.}
  \label{tab:ablation-aware}
  \begin{threeparttable}
  \begin{tabular}{lcccccc}
    \toprule
    Variant            & SE & SM & SH & HE & HM & HH  \\
    \midrule
    FAR-AVIO (full)          & 0.11 & 0.16 & 0.13 & 0.19 & 0.33 & 0.25 \\
    FAR-AIO  & 0.24 & 0.21 & 0.14 & 0.39 & F\tnote{1} & F \\
    FAR-AVIO w/o AWARE       & 0.13 & F & F & F & F & F \\
    \bottomrule
  \end{tabular}
  \begin{tablenotes}
    \small
    \item[1] F mean tracking failure or filter divergence.
  \end{tablenotes}
  \end{threeparttable}
\end{table}

\paragraph{Effect of IMU--DVL extrinsic calibration}
\begin{table}[t]
  \centering
  \caption{Ablation of the extrinsic calibration module on Tank and simulate sequences.}
  \label{tab:ablation-calib}
  \begin{threeparttable}
  \begin{tabular}{lcccccc}
    \toprule
    Variant             & SE & SH & Identity & NM\tnote{2} & NS\tnote{2} & Average \\
    \midrule
    FAR-AIO    & 0.242 & 0.212 & 0.574 & 0.165 & 0.124 & \textbf{0.263} \\
    w/o calib\tnote{1}  & 0.323 & 0.217 & 9.372 & 3.546 & 2.848 & 8.152 \\
    \bottomrule
  \end{tabular}
  \begin{tablenotes}
    \small
    \item[1] w/o calib mean FAR-AIO without extrinsic calibration.
    \item[2] NM mean middle level noise; NS mean small level noise.
  \end{tablenotes}
  \end{threeparttable}\vspace{-2mm}
\end{table}

We first validate the IMU–DVL extrinsic calibration convergence in a numerical simulation. Figure~\ref{fig:dvl-extrinsic-calib} plots the evolution of the extrinsic errors over time for three different initializations: an identity transform (\textit{Identity}), and two perturbed initial extrinsics with medium (\textit{Noise Middle}) and small (\textit{Noise Small}) noise levels. In all cases, the estimated IMU–DVL extrinsics converge toward the ground truth, with the final errors stabilizing at a small residual level. Larger initial perturbations understandably lead to longer convergence transients and slightly higher steady-state error, but the calibration remains stable and convergent even from the coarse identity initialization. This demonstrates that the proposed calibration scheme can reliably recover extrinsics from IMU–DVL motion without requiring a carefully tuned initial guess.

We further quantify the impact on trajectory accuracy, as shown in Table~\ref{tab:ablation-calib}., both real Tank sequences and simulated sequences are initialized with perturbed extrinsics. On Tank data, calibration reduces RMSE by 10--25\%, in simulation with larger perturbations, corrupted extrinsics cause RMSE to exceed 3--9\,m without calibration, whereas online calibration recovers sub-meter accuracy ($0.124$--$0.574$\,m). On average, enabling calibration reduces RMSE from $8.152$\,m to $0.263$\,m, confirming that the proposed module reliably recovers accurate extrinsics from noisy initial guesses.

\section{Conclusions}
\label{sec:conclusion}

This paper presented FAR-AVIO, a fast and robust Schur-Complement based Acoustic-Visual-Inertial fusion odometry framework with online sensor calibration for underwater robots. Extensive evaluations on real world sequences and synthetic scenarios demonstrated that FAR-AVIO achieves competitive or superior localization accuracy compared with state-of-the-art underwater and terrestrial baselines, while requiring substantially lower CPU and memory resources and running comfortably in real time on embedded hardware.




\bibliography{misc/ref.bib}

\begin{thebibliography}{10}
\providecommand{\url}[1]{#1}
\csname url@samestyle\endcsname
\providecommand{\newblock}{\relax}
\providecommand{\bibinfo}[2]{#2}
\providecommand{\BIBentrySTDinterwordspacing}{\spaceskip=0pt\relax}
\providecommand{\BIBentryALTinterwordstretchfactor}{4}
\providecommand{\BIBentryALTinterwordspacing}{\spaceskip=\fontdimen2\font plus
\BIBentryALTinterwordstretchfactor\fontdimen3\font minus \fontdimen4\font\relax}
\providecommand{\BIBforeignlanguage}[2]{{%
\expandafter\ifx\csname l@#1\endcsname\relax
\typeout{** WARNING: IEEEtran.bst: No hyphenation pattern has been}%
\typeout{** loaded for the language `#1'. Using the pattern for}%
\typeout{** the default language instead.}%
\else
\language=\csname l@#1\endcsname
\fi
#2}}
\providecommand{\BIBdecl}{\relax}
\BIBdecl

\bibitem{aqualocdataset}
M.~Ferrera, V.~Creuze, J.~Moras, and P.~Trouv\'{e}-Peloux, ``Aqualoc: An underwater dataset for visual–inertial–pressure localization,'' \emph{Int. J. Robot. Research}, vol.~38, no.~14, pp. 1549--1559, 2019.

\bibitem{TankDatasetUnderwaterxu2025a}
S.~Xu, J.~Scharff~Willners, J.~Roe, S.~Katagiri, T.~Luczynski, Y.~Petillot, and S.~Wang, ``Tank dataset: {{An}} underwater multi-sensor dataset for {{SLAM}} evaluation,'' \emph{Int. J. Robot. Research}, 2025.

\bibitem{campos2021orb}
C.~Campos, R.~Elvira, J.~J.~G. Rodr{\'\i}guez, J.~M. Montiel, and J.~D. Tard{\'o}s, ``{ORB-SLAM3}: An accurate open-source library for visual, visual--inertial, and multimap {SLAM},'' \emph{{IEEE} Trans. Robot.}, vol.~37, no.~6, pp. 1874--1890, 2021.

\bibitem{qin2018vins}
T.~Qin, P.~Li, and S.~Shen, ``{VINS-Mono}: A robust and versatile monocular visual-inertial state estimator,'' \emph{{IEEE} Trans. Robot.}, vol.~34, no.~4, pp. 1004--1020, 2018.

\bibitem{dmvio}
L.~v. Stumberg and D.~Cremers, ``{DM-VIO}: Delayed marginalization visual-inertial odometry,'' \emph{{IEEE} Robotics and Automation Letters}, vol.~7, no.~2, pp. 1408--1415, 2022.

\bibitem{ov2}
M.~Ferrera, A.~Eudes, J.~Moras, M.~Sanfourche, and G.~Le~Besnerais, ``{OV$^{2}$SLAM}: A fully online and versatile visual {SLAM} for real-time applications,'' \emph{IEEE Robotics and Automation Letters}, vol.~6, no.~2, pp. 1399--1406, Apr. 2021.

\bibitem{TightlyCoupledVisualInertialPressureFusionhu2022a}
C.~Hu, S.~Zhu, Y.~Liang, and W.~Song, ``Tightly-{{Coupled Visual-Inertial-Pressure Fusion Using Forward}} and {{Backward IMU Preintegration}},'' \emph{IEEE Robotics and Automation Letters}, vol.~7, no.~3, pp. 6790--6797, Jul. 2022.

\bibitem{vip-init}
C.~Hu, S.~Zhu, Y.~Liang, Z.~Mu, and W.~Song, ``Visual-pressure fusion for underwater robot localization with online initialization,'' \emph{{IEEE} Robotics and Automation Letters}, vol.~6, no.~4, pp. 8426--8433, 2021.

\bibitem{AQUASLAMTightlyCoupledxu2025}
S.~Xu, K.~Zhang, and S.~Wang, ``{{AQUA-SLAM}}: Tightly coupled underwater acoustic-visual-inertial slam with sensor calibration,'' \emph{{IEEE} Trans. Robot.}, vol.~41, pp. 2785--2803, 2025.

\bibitem{SVIn2MultisensorFusionbasedrahman2022a}
S.~Rahman, A.~Quattrini~Li, and I.~Rekleitis, ``{{SVIn2}}: {{A}} multi-sensor fusion-based underwater {{SLAM}} system,'' \emph{Int. J. Robot. Research}, vol.~41, no. 11-12, pp. 1022--1042, Sep. 2022.

\bibitem{TightlyCoupledVisualDVLzhao2023}
L.~Zhao, M.~Zhou, and B.~Loose, ``Tightly coupled visual-dvl-inertial odometry for robot-based ice-water boundary exploration,'' in \emph{Proc. of the {IEEE/RSJ} Int. Conf. on Intell. Robots and Syst.}, 2023, pp. 7127--7134.

\bibitem{WangsenUnderwaterVisualAcousticSLAM2021}
E.~Vargas, R.~Scona, J.~S. Willners, T.~Luczynski, Y.~Cao, S.~Wang, and Y.~R. Petillot, ``Robust underwater visual slam fusing acoustic sensing,'' in \emph{Proc. of the {IEEE} Int. Conf. on Robot. and Autom.}, May 2021, pp. 2140--2146.

\bibitem{SlidingWindowFiltersibley2010}
G.~Sibley, L.~Matthies, and G.~Sukhatme, ``Sliding window filter with application to planetary landing,'' \emph{J. Field Robot.}, vol.~27, no.~5, pp. 587--608, Sep. 2010.

\bibitem{fan2024schurvins}
Y.~Fan, T.~Zhao, and G.~Wang, ``{SchurVINS}: Schur complement-based lightweight visual inertial navigation system,'' in \emph{Proc. of the {IEEE} Int. Conf. on Pattern Recognition}, 2024, pp. 17\,964--17\,973.

\bibitem{geneva2020openvins}
P.~Geneva, K.~Eckenhoff, W.~Lee, Y.~Yang, and G.~Huang, ``Openvins: A research platform for visual-inertial estimation,'' in \emph{Proc. of the {IEEE} Int. Conf. on Robot. and Autom.}, 2020, pp. 4666--4672.

\bibitem{kumar-msckf}
K.~Sun, K.~Mohta, B.~Pfrommer, M.~Watterson, S.~Liu, Y.~Mulgaonkar, C.~J. Taylor, and V.~Kumar, ``Robust stereo visual inertial odometry for fast autonomous flight,'' \emph{{IEEE} Robotics and Automation Letters}, vol.~3, no.~2, pp. 965--972, 2018.

\bibitem{svo}
C.~Forster, M.~Pizzoli, and D.~Scaramuzza, ``{{SVO}}: {{Fast}} semi-direct monocular visual odometry,'' in \emph{Proc. of the {IEEE} Int. Conf. on Robot. and Autom.}, May 2014, pp. 15--22.

\bibitem{okvis2014}
S.~Leutenegger, S.~Lynen, M.~Bosse, R.~Siegwart, and P.~Furgale, ``Keyframe-based visual--inertial odometry using nonlinear optimization,'' \emph{Int. J. Robot. Research}, vol.~34, no.~3, pp. 314--334, 2015.

\bibitem{xushida-underwater-visual-acoustic-slam-with-extrinsic-calibration-2021}
S.~Xu, T.~Luczynski, J.~S. Willners, Z.~Hong, K.~Zhang, Y.~R. Petillot, and S.~Wang, ``Underwater visual acoustic {SLAM} with extrinsic calibration,'' in \emph{Proc. of the {IEEE/RSJ} Int. Conf. on Intell. Robots and Syst.}, 2021, pp. 7647--7652.

\bibitem{Visual-DVL-Fusion-HuangyuPei}
Y.~Huang, P.~Li, S.~Yan, Y.~Ou, Z.~Wu, M.~Tan, and J.~Yu, ``Tightly-coupled visual-dvl fusion for accurate localization of underwater robots,'' in \emph{Proc. of the {IEEE/RSJ} Int. Conf. on Intell. Robots and Syst.}, 2023, pp. 8090--8095.

\bibitem{lidar-dvl-fusion-thoms}
A.~Thoms, G.~Earle, N.~Charron, and S.~Narasimhan, ``Tightly coupled, graph-based dvl/imu fusion and decoupled mapping for {SLAM}-{Centric} maritime infrastructure inspection,'' \emph{{IEEE} J. Oceanic Eng.}, vol.~48, no.~3, pp. 663--676, 2023.

\bibitem{HuangYupei-TC}
Y.~Huang, P.~Li, S.~Ma, S.~Yan, M.~Tan, J.~Yu, and Z.~Wu, ``Visual-inertial-acoustic sensor fusion for accurate autonomous localization of underwater vehicles,'' \emph{{IEEE} Trans. Cybernetics}, vol.~55, no.~2, pp. 880--896, 2025.

\bibitem{eskf-solar}
\BIBentryALTinterwordspacing
J.~Sol{\`{a}}, ``Quaternion kinematics for the error-state kalman filter,'' \emph{CoRR}, vol. abs/1711.02508, 2017. [Online]. Available: \url{http://arxiv.org/abs/1711.02508}
\BIBentrySTDinterwordspacing

\bibitem{ShiTomasi}
J.~Shi and C.~Tomasi, ``Good features to track,'' in \emph{Proc. of the {IEEE} Int. Conf. on Pattern Recognition}, 1994, pp. 593--600.

\bibitem{Lucas-optical-flow}
B.~D. Lucas and T.~Kanade, ``An iterative image registration technique with an application to stereo vision,'' in \emph{Proc. Int. Joint Conf. Artif. Intell.}, 1981, pp. 674--679.

\bibitem{Umeyama}
S.~Umeyama, ``Least-squares estimation of transformation parameters between two point patterns,'' \emph{{IEEE} Trans. Pattern Anal. and Mach. Intell.}, vol.~13, no.~4, pp. 376--380, 1991.

\end{thebibliography}

\end{document}